\pdfoutput=1

\documentclass[11pt]{article}

\usepackage{acl}

\usepackage{times}
\usepackage{latexsym}
\usepackage{graphicx}
\usepackage{amsmath}
\usepackage[T1]{fontenc}
\usepackage{booktabs}
\usepackage{multirow}
\usepackage{multicol}
\usepackage{graphicx}
\usepackage{subcaption}
\usepackage[utf8]{inputenc}

\usepackage{microtype}

\usepackage{inconsolata}

\usepackage{xcolor}
\definecolor{myp}{RGB}{128, 0, 128}
\newcommand{\name}{\textit{HoPE}}
\title{\name: A Novel Positional Encoding Without Long-Term Decay for Enhanced Context Awareness and Extrapolation}

\author{
\begin{tabular}{c}
Yuhan Chen$^{1}$ \quad  Ang\  Lv$^{2}$  \quad \textbf{Jian Luan}$^{1}$ \quad  \textbf{Bin Wang}$^{1}$ \quad \textbf{Wei Liu}$^{1}$\thanks{\ \ Corresponding author.}
\end{tabular}
\\ \vspace{.5mm}
    \small
    \begin{tabular}{c}
    $^1$Xiaomi AI Lab \quad $^2$Gaoling School of Artificial Intelligence, Renmin University of China\\
    \end{tabular}
    \\ \vspace{.5mm}
    \small
    \begin{tabular}{c}
    \texttt{\{chenyuhan5, luanjian, wangbin11, liuwei40\}@xiaomi.com}\\
    \texttt{anglv@ruc.edu.cn} \\
    \end{tabular}
    \vspace{2mm} \\
}

\begin{document}
\maketitle
\begin{abstract}
Many positional encodings (PEs) are designed to exhibit long-term decay, based on an entrenched and long-standing inductive opinion: tokens farther away from the current position carry less relevant information. 
We argue that long-term decay is outdated in the era of LLMs, as LLMs are now applied to tasks demanding precise retrieval of in-context information from arbitrary positions.
Firstly, we present empirical analyses on various PEs, demonstrating that models inherently learn attention with only a local-decay pattern while forming a U-shape pattern globally, contradicting the principle of long-term decay. 
Furthermore, we conduct a detailed analysis of rotary position encoding (RoPE, a prevalent relative positional encoding in LLMs), and found that the U-shape attention is caused by some learned components, which are also the key factor limiting RoPE's expressiveness and extrapolation.
Inspired by these insights, we propose \textbf{H}igh-frequency r\textbf{o}tary \textbf{P}osition \textbf{E}ncoding (\name).
\name\ replaces the specific components in RoPE with position-independent ones, retaining only high-frequency signals, which also breaks the principle of long-term decay in theory.  \name\ achieves two major advantages: 
(1) Without constraints imposed by long-term decay, contradictory factors that limit attention optimization are removed. Thus, the model's context awareness is enhanced.
(2) \name\ exhibits greater robustness to the out-of-distribution behavior in attention patterns during extrapolation. 
The effectiveness of \name\ is validated through extensive experiments and with a large language model of up to \textit{3 billion} parameters.

\end{abstract}

\section{Introduction}
\label{sec:intro}

\begin{table}[t]
\centering
\begin{tabular}{lcc}
\toprule
Benchmark & RoPE  & \name\ (Ours)  \\ \hline
MMLU      & 34.27 & \textbf{38.38} \\
MMLU-PRO  & 12.60  & \textbf{12.74} \\
GPQA      & 23.23 & \textbf{28.28} \\
BBH       & 29.00 & \textbf{29.15} \\
WinoGrande & \textbf{51.70} & 50.43 \\
GSM8k     & 10.61 & \textbf{12.05} \\
MATH      & 1.16  & \textbf{1.84}  \\
DROP      & 31.29 & \textbf{38.46} \\\hline
AVG.       & 24.23 & \textbf{26.42} \\
\bottomrule
\end{tabular}
\caption{Performance comparison between RoPE and \name\ across eight benchmarks using a 3B Llama-based model trained with a sequence length of 8192 over 500 billion tokens. Better results are highlighted in \textbf{bold fonts}. \name\ demonstrates superior performance in most tasks.}
\label{tab:3b}
\end{table}

Positional encoding (PE) plays a crucial role in Transformers~\cite{Vaswani2017AttentionIA} to capture the order of input sequence, as the attention mechanism is permutation invariant. The original PE proposed by \citet{Vaswani2017AttentionIA} struggles to generalize beyond the training sequence length. To address this limitation, relative positional encoding (RPE) methods have been introduced, including RoPE~\cite{Su2021RoFormerET}, ALiBi~\cite{Press2021TrainST}, and KERPLE~\cite{Chi2022KERPLEKR}.
These RPEs share a long-standing and entrenched design
~\cite{Su2021RoFormerET}: the long-term decay, i.e., tokens with a long relative distance should receive less attention. 

However, in the era of LLMs, a question arises: is it still necessary to retain this design? As LLMs are increasingly used in long-text scenarios—where they must leverage distant information, such as in Retrieval-Augmented Generation (RAG,~\citealp{Izacard2020LeveragingPR})—long-term decay potentially limits their performance.

In this paper, we demonstrate that the answer to the above question is no.
Through empirical analyses on various PEs, we found that the attention patterns learned by models tend to contradict the principle of long-term decay. Specifically, models only retain a local-decay pattern, while learning a U-shape attention distribution globally. 
We further delve into an analysis of RoPE (Rotary Position Encoding by~\citealp{Su2021RoFormerET}, a widely-used RPE), which claims to ensure the long-term decay by combining various frequency components (See Section~\ref{sec:relate} for details) while empirically learns the U-shape pattern. 
We decomposed these components and obtained the following observations:

(1) In RoPE, some components of specific frequency (which we named ``activated'' components in this paper) play a key role in shaping the final U-shape attention pattern and they exhibit fluctuations similar to the overall attention pattern. 
We observed that these components have a predominant impact on attention in the early stages of training.
As the training steps increase, however, the model attempts to offset their effects, increasing the weight of other components. 
We assume this reflects a form of shortcut learning behavior~\cite{Geirhos2020ShortcutLI, Robinson2021CanCL, Du2022ShortcutLO}, which hinders the optimization. 
We further found that the frequency of activated components can be pre-calculated before training, based on the relationship between the RoPE rotary angle and the training context length.

(2) We explored the attention patterns in extrapolation tasks and found that ``activated'' components are a key factor limiting RoPE’s extrapolation abilities. These components cause out-of-distribution (OOD) attention logits in the first layer during extrapolation, triggering a cascade of disrupted attention patterns through the subsequent layers.

(3) The top low-frequency components (whose frequency is lower than the ``activated'' components) tend to stabilize as constant patterns, with a small magnitude. This indicates that these components are not being effectively utilized for representing positional information and learn more about semantics information.

Based on the findings above, we summarize three key insights: (1) Global long-term decay is not necessary for the model and may even hinder optimal learning. (2) To enhance the model's context awareness and extrapolation, the frequencies of RoPE's learned components should be constrained. 
(3) There is redundancy in RoPE, and representation subspaces occupied by certain components could be better utilized.

In this paper, we propose a novel positional encoding method called \textbf{H}igh-frequency r\textbf{o}tary \textbf{P}osition \textbf{E}ncoding (\name). 
\name\ follows above insights and is quite intuitive to implement: we replace the ``doomed-to-be-activated'' and top low-frequency components in the original RoPE with position-independent ones, while retaining the high-frequency components.
As a result, contradictory factors for attention optimization are eliminated, extrapolation limitations are reduced, and position information is still well-represented by high-frequency signals.

On small language models with 125 million parameters, we assess the model's potential both within and beyond the context length by evaluating perplexity, in-context copying ability and few-shot following ability.
\name\ demonstrates superior performance compared to other PEs.
We further trained large language models with 3 billion parameters from scratch, and found \name\ performs better than RoPE in complex NLP tasks.

To sum up, we make three major contributions:

(1) We show that long-term decay in PEs is unnecessary in the era of large models, as supported by empirical analysis of various PEs. 

(2) We explore the relationship between the overall attention pattern and the decomposed components of RoPE, and propose a new explanation for RoPE's limited performance and poor extrapolation.

(3)  Based on the above insights, we design \name, a novel relative positional encoding. Experiments empirically validate the effectiveness of \name.

\section{Related Work}
\label{sec:relate}
Positional encoding is a fundamental component of Transformer models~\cite{Vaswani2017AttentionIA}, addressing the lack of sequential information inherent in self-attention mechanisms. The existing positional embeddings can be broadly categorized into absolute (APE) and relative (RPE) various.
\subsection{Absolute positional encoding (APE)}
Absolute positional encoding~\cite{Vaswani2017AttentionIA, Wang2021OnPE, Kiyono2021SHAPESA} focuses solely on individual position information and is typically applied in the first layer of the model. It is implemented by assigning a (learnable or fixed sinusoidal) real-valued encoding ${pe}_{i}$ to each position $i$ and is integrated into the representation of input sequences through simple vector addition. 

\subsection{Relative positional encoding (RPE)}
Although absolute positional encoding (APE) is simple and intuitive, it struggles to generalize effectively for long sequences. As a result, recent research has focused primarily on optimizing relative positional encoding (RPE)~\cite{Shaw2018SelfAttentionWR, Raffel2019ExploringTL, Lv2023AreWF}. Currently, the popular RPE methods can be divided into two main types~\cite{Zheng2024DAPEDP}: rotary position encoding and addition position encoding.

\textbf{Rotary position encoding} (RoPE, \citealp{Su2021RoFormerET}) encodes positional information by rotating the query and key vectors. 
In each Transformer layer, RoPE applies a $d$-dimensional rotation matrix (denoted as $R_{\theta,m}$) to the query or key vector at position $m$ in the sequence for positional encoding. The specific inner product process can be illustrated as follows: 
\begin{equation}
\begin{aligned}
&q_{m}=R_{\Theta, m} W_{q} x_{m} =R_{\Theta, m}q,\\
&k_{n}=R_{\Theta, n} W_{k} x_{n}=R_{\Theta, n}k,\\
q_{m} \cdot k_{n} &= (R_{\Theta, m} q)^{\top} (R_{\Theta, n} k) = q^{\top} R_{\Theta, m-n} k 
\end{aligned}
\label{eq:re}
\end{equation}

where $x$ is the $d$-dimensional input of the current Transformer layer, and the matrix $R_{\Theta,m}$ is a block diagonal matrix consisting of $d/2$ blocks, each of which size $2 \times 2$ and assigned a specific angle $\theta$. This is defined as:
\begin{equation}
\begin{aligned}
&R_{\theta_i, m} = \begin{bmatrix}
    \cos(m \theta_{i}) & -\sin(m \theta_{i}) \\
    \sin(m \theta_{i}) & \cos(m \theta_{i})
\end{bmatrix}, \\
&R_{\Theta, m} = \texttt{Diag}( R_{\theta_0, m},  ...,  R_{\theta_{d/2-1}, m} )
\end{aligned}
\label{eq:rope}
\end{equation}
where $\theta_i = {b}^{-\frac{2i}{d}}$, and $b$ is referred to as the base of the rotary angle.

This encoding method cleverly computes the inner product of relative positions by encoding absolute positions without altering the attention computation process, making it more compatible with various efficient inference methods. However, the original RoPE encoding exhibits poor extrapolation capability for longer sequences~\cite{Press2021TrainST, Kazemnejad2023TheIO}. This raises one popular research direction for exploring RoPE-based length extrapolation methods, such as PI~\cite{Chen2023ExtendingCW},  LongRoPE~\cite{Ding2024LongRoPEEL}, Randomized RoPE~\cite{Ruoss2023RandomizedPE} and YaRN~\cite{Peng2023YaRNEC}.

\textbf{Additive relative positional encoding} (ARPE) is another popular method, which introduces a bias matrix $B$ to the original (pre-softmax) attention logits. This approach can be uniformly formula as follows.
\begin{equation}
\begin{aligned}
    {Attn}_{ARPE}(X) =  XW_{Q}(XW_{K})^{T} + B
\end{aligned}
\end{equation}
Different designs of the bias matrix $B$ result in various APE variants, including T5’s Bias~\cite{Raffel2019ExploringTL}, ALiBi~\cite{Press2021TrainST}, KERPLE~\cite{Chi2022KERPLEKR}, Sandwich~\cite{Chi2022DissectingTL}, and FIRE~\cite{Li2024FunctionalIF}. These ARPE methods claim robust performance in length extrapolation, as measured by the perplexity (PPL).  Nevertheless, some studies~\cite{Press2021TrainST} noted that PPL may not accurately represent real task performance. Our study further confirms that some ARPEs fail to effectively leverage global information, resulting in only marginal improvements in actual length extrapolation.

\section{Discussion on Position-related Attention Pattern}
In this section, We first present the position-related attention patterns (within the training context length) learned by three PEs. We observed that, although the long-term decay of PEs is intuitive, this decay is not global in the empirical attention patterns. Instead, the attention patterns tend to resemble a U-shape curve.

Secondly, we delve into a detailed analysis of the relationship between this U-shape pattern and the various components (assigned with different frequencies) of RoPE. 
We found that the overall pattern is strongly correlated with some components with specific frequencies, which are key factors to limit model's context awareness and extrapolation.

\subsection{Experiments Setups}
\label{sec:setting}

We train small Llama language models~\cite{Touvron2023LLaMAOA, Touvron2023Llama2O, Dubey2024TheL3} with 125 million parameters, using different PEs. 
The training dataset contains 200 billion tokens sourced from RedPajama~\cite{together2023redpajama}. All experiments use Llama tokenizer with a vocabulary of 32,000 tokens. The training context length is 512 tokens and the update steps are 50,000.
Detailed configurations and other hyperparameters are provided in the Appendix~\ref{app:setting}.

To observe the position-related attention patterns both within and beyond the training context length, we set two test lengths: 512 and 1024. For each test length, we generate 5,000 corresponding data samples, each assigning a random token from the vocabulary to all positions in the input sequence (except for the initial [bos]). We then calculate the pre-softmax attention logits for each position. To illustrate a common pattern, we average the results across all layers and heads, as most heads demonstrate similar behaviors.

\subsection{Long-term Decay in Attention Patterns}

\label{sec:pattern-all}

Our analysis focused on three PEs including learnable APE, RoPE, and KERPLE. We don't take ALiBi into account, as its bias matrix $B$ is unlearnable and forces the attention pattern to be global long-term decay.

\begin{figure}[t]
    \centering
    \includegraphics[width=\linewidth,height=6cm]{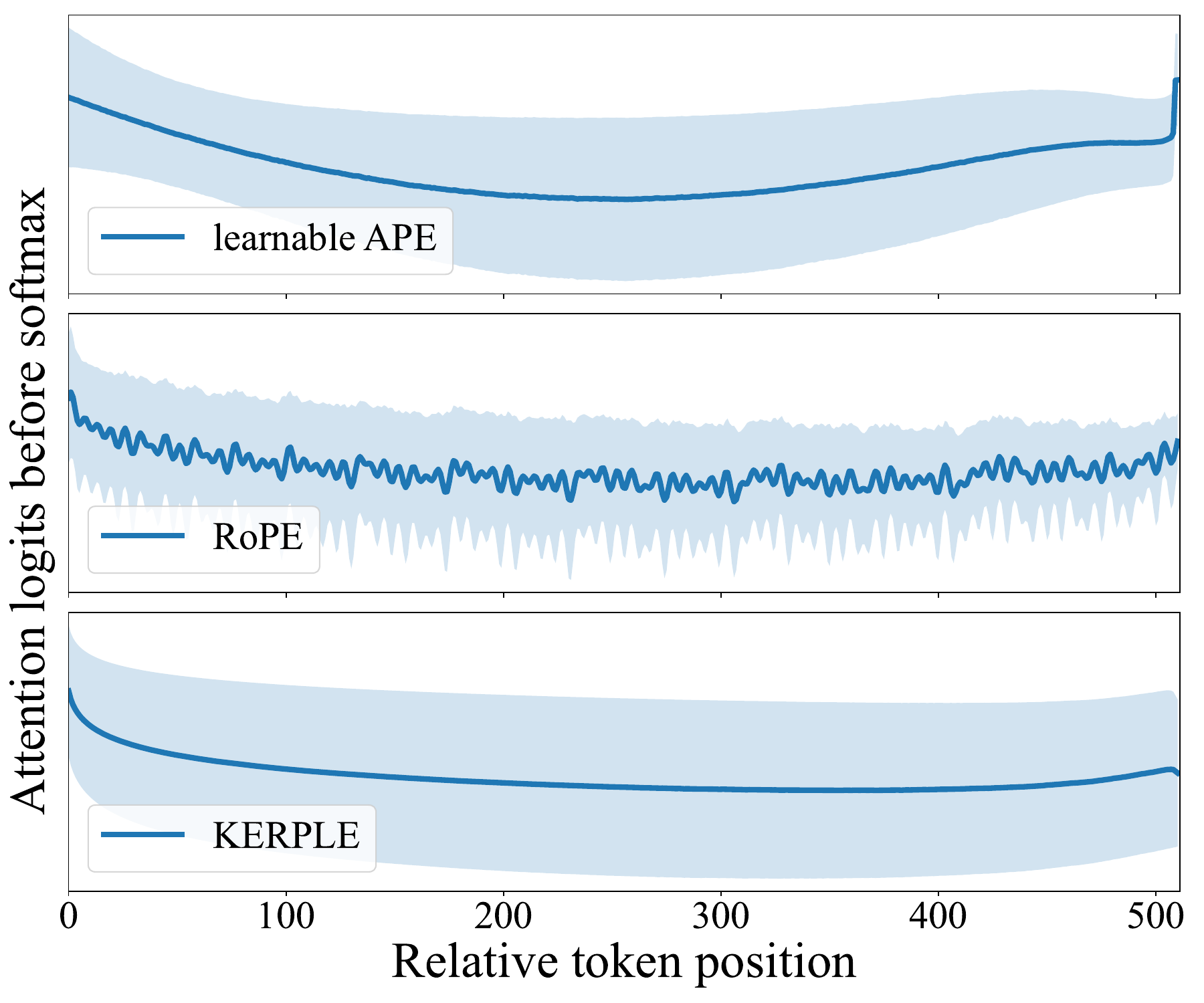}
    \caption{Attention patterns based on different learnable position encodings: (a) learnable APE, (b) RoPE, (c) KERPLE. }
    \label{fig:pattern-all}
\end{figure}

 Results are shown in Figure~\ref{fig:pattern-all}.
 One important observation is that the attention patterns \textbf{do not exhibit global long-term decay}. Instead, the attention patterns tend to form a U-shape curve, which ensures the decay of adjacent tokens while increasing the importance of the initial tokens.

\begin{figure*}[t]
    \centering
    \begin{subfigure}[b]{0.32\textwidth}
    \includegraphics[width=\linewidth]{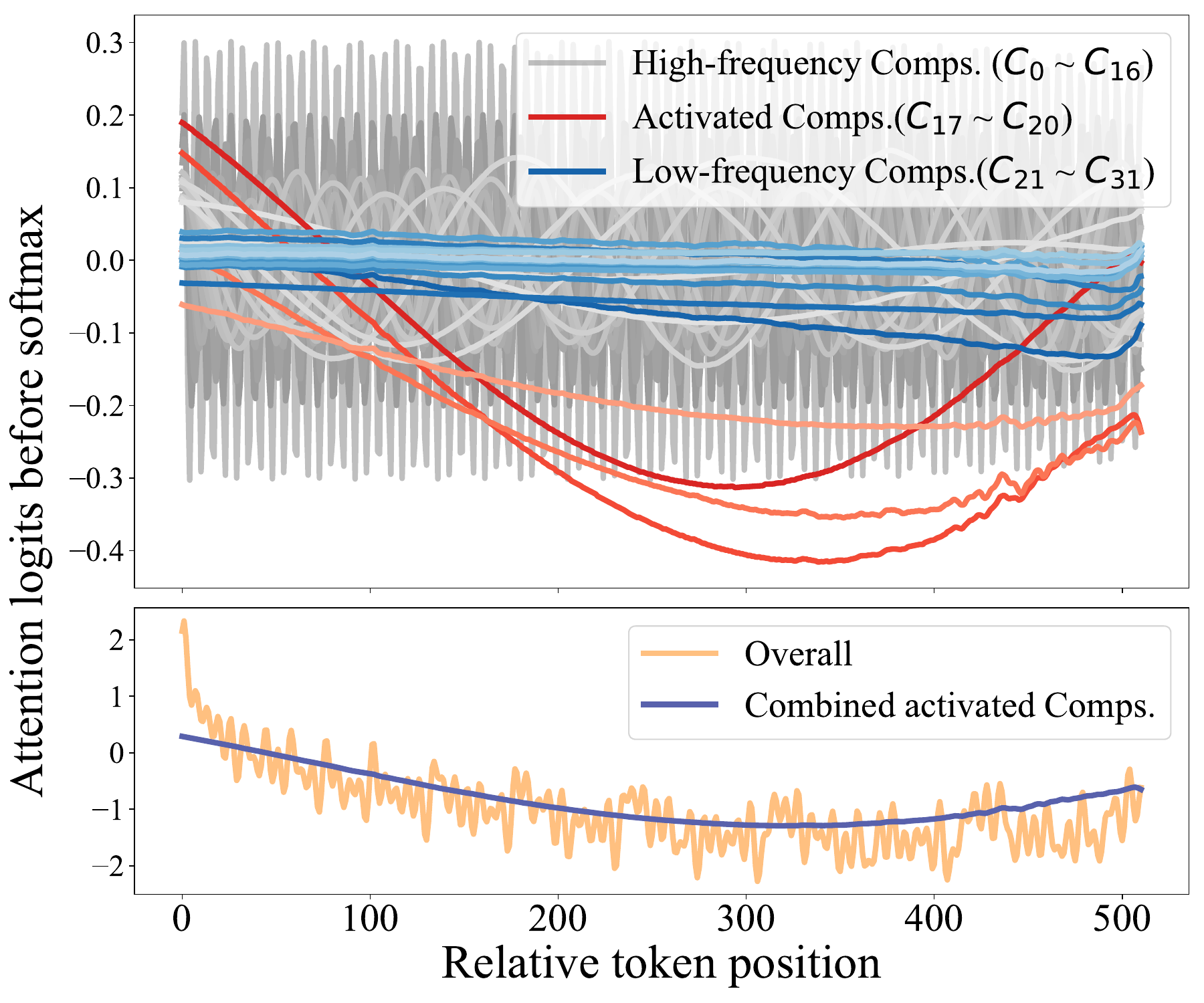}
    \caption{}
    \label{fig:pattern-rope-b}
    \end{subfigure}
    \begin{subfigure}[b]{0.32\textwidth}
    \includegraphics[width=\linewidth]{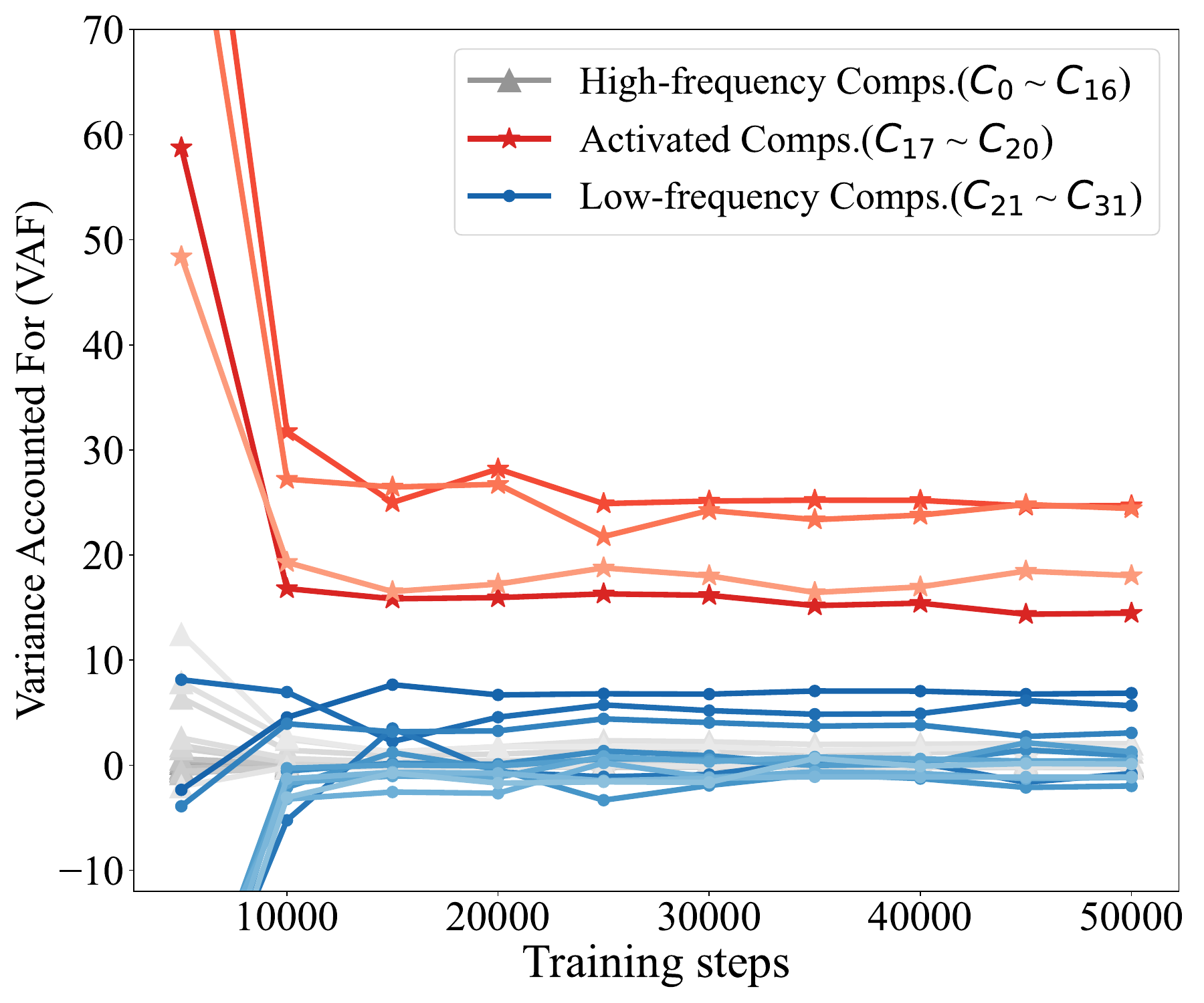}
    \caption{}
    \label{fig:pattern-rope-a}
    \end{subfigure}
    \begin{subfigure}[b]{0.32\textwidth}
    \includegraphics[width=\linewidth]{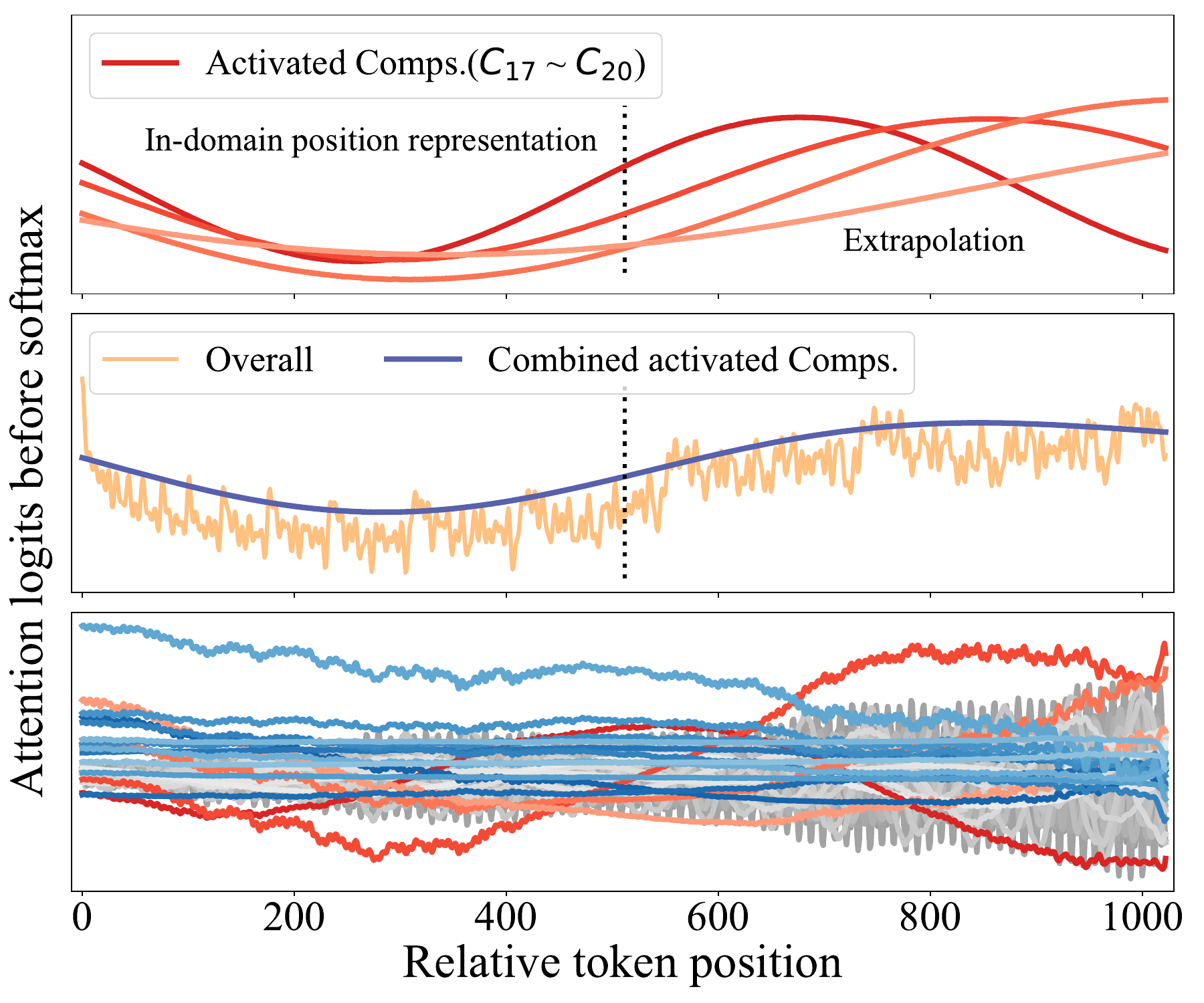}
    \caption{}
    \label{fig:pattern-rope-c}
    \end{subfigure}

    \caption{
    (a) We decomposed RoPE into components (Comps.) for analysis (See Eq.\ref{eq:comp}).
    The upper subplot displays the contribution to the overall attention logits from each component.
    We highlight some components with outstanding patterns, namely ``activated'' components, in red and lower frequency components in blue.
    The lower subplot presents the overall attention logits, along with the combined effects of ``activated'' components we highlighted in the top figure.
    (b) Variance Accounted For (VAF, See Section~\ref{para:vaf}) for different components of RoPE during training.
    (c) The OOD phenomenon in extrapolation caused by ``activated'' components. The two upper subplots show the attention patterns of the first layer, and the lower subplot presents the anomalous patterns of the subsequent layers.
    The model training length presented here is 512, results for training length in 1,024 can be found in Appendix~\ref{app:pattern-rope}.
    }
    \label{fig:pattern-rope}
\end{figure*}

\subsection{Effects of Different Components of RoPE in Attention Pattern}
\label{para:rope-result}

Figure~\ref{fig:pattern-all} also demonstrated that RoPE empirically learns the U-shape pattern while claiming to employ multiple components with different frequencies to ensure long-term decay attention.  
We wonder which components truly matter in this process and delve into a detailed analysis.

\subsubsection{Preliminaries}
 \label{para:rope}
 \paragraph{Components of RoPE} According to Eq.\ref{eq:rope}, we can see that the dot product in attention can be broken down into an inner product process of d/2 components, each with a distinct angle $\theta_i$, followed by a summation. This can be expressed by the following formula, which allows us to explore the individual effect of each positional component $C_i$.
\begin{equation}
\begin{aligned}
& q_{m} \cdot k_{n} = q^{T} R_{\Theta, m-n} k = \sum_{i=0}^{d/2-1}\underbrace{{q_{i}^{T} R_{\theta_i, m-n} k_{i}}}_{C_i} \\
&= \sum_{i=0}^{d/2-1}((q_{i,0}k_{i,0} + q_{i,1} k_{i,1})\cdot cos((m-n)\theta_i)) \\
&+ ( q_{i,0}k_{i,1} - q_{i,1} k_{i,0})\cdot sin((m-n)\theta_i))
\end{aligned}
\label{eq:comp}
\end{equation}

\paragraph{Variance Accounted For (VAF)}
\label{para:vaf}
VAF~\cite{Yoon2021RegularizedNR, Qiu2021PerformanceEO} is primarily used to measure the explanatory power of components for the total variability. It serves as a crucial criterion for identifying effective principal components. A larger value indicates that the component holds greater importance. The formula is as follows: 
\begin{equation}
{VAF}_{\hat{y},y}(\%) = [ 1 - \frac{\sum_{i=1}^n(y_i - \hat{y}_i)}{\sum_{i=1}^ny_i} ] \times 100
\end{equation}
where $\hat{y}$ is a component of $y$.

 \subsubsection{Experiments and Results}

We decompose RoPE into several components, each associated with a unique frequency \(\theta\), and examine their individual contribution to the overall attention logits, both within the training length and during extrapolation.  
Additionally, under the scenarios within training context length, we tracked the pattern dynamics of components as training progresses, using VAF metric.

The results are presented in Figure~\ref{fig:pattern-rope}.\footnote{To further validate our findings, we also included results with a longer training context length of 1,024 in Appendix~\ref{app:pattern-rope}.} From these results, we can derive three key insights.
 
 \textbf{(1) The learning of attention patterns is closely associated with some specific components in RoPE, while the model tends to counteract these ``activated'' components during training.} As seen in Figure~\ref{fig:pattern-rope-a}, some components (referred to as ``activated'' components, highlighted with red in Figure~\ref{fig:pattern-rope-a}) exhibit high VAF, indicating that they dominate the formation of the overall U-shape pattern. The lower subplot in Figure~\ref{fig:pattern-rope-b} further confirms this, as the combined pattern of these components mirrors the fluctuations of the overall pattern. However, as indicated in Figure~\ref{fig:pattern-rope-a}, the VAF values of the ``activated'' components decrease as training progresses, suggesting that the model is reducing the contribution of these components. We consider this phenomenon a form of shortcut learning~\cite{Geirhos2020ShortcutLI, Robinson2021CanCL, Du2022ShortcutLO}, which may constrain the model's overall learning. We also found that all these ``activated'' components exhibit U-shaped fluctuations across varying training lengths (as seen in the upper subplot of Figure~\ref{fig:pattern-rope-b}). Upon further examination of these components, we found that their frequencies ($\theta$s) fall within the range $ \left( \frac{\pi}{L},  \frac{2\pi}{L} \right)$,
where $L$ is the training length.

\textbf{(2) The ``activated'' components in RoPE are also the primary reason for its poor extrapolation ability.}
As mentioned above, the attention pattern is closely tied to the ``activated'' components, which
exhibit U-shape (or low half-cycle) patterns with the training length. Considering the cosine properties of these components (see Section~\ref{para:rope} for detail), we can clearly observe from the attention pattern in the first layer (shown in Figure~\ref{fig:pattern-rope-c}) that these ``activated'' components are located in the upper half-cycle when extrapolation, which is a significant out-of-distribution (OOD) phenomenon, and subsequently leads to the disarray of attention patterns in later layers.

\textbf{(3) Components with a lower frequency than the ``activated'' ones tend to learn a constant pattern and are not effectively utilized.}
Another observation is that many components exhibit a constant pattern despite their cosine properties, as shown in the upper subplot of Figure~\ref{fig:pattern-rope-b}. Upon delving deeper into these components, we found that their frequencies are all lower than the ``activated'' components. We speculate that these top low-frequency components do not represent positional information, but rather semantic information. And the properties constraints on them may even hinder this learning, as the corresponding patterns exhibit small magnitudes.

\section{A Novel PE Enhances Model's Context Awareness and Exploration }

Inspired by all experimental results and observations above, we proposed \textbf{H}igh-frequency r\textbf{o}tary \textbf{P}osition \textbf{E}ncoding (\name). With slight modification in RoPE, \name\ greatly improves the model's context awareness and extrapolation. We first detail our approach and then validate its effectiveness on perplexity, copying task, and few-shot following tasks. The results demonstrate that \name\ exhibits superior performance compared to other PEs.

\subsection{Method}
We propose our method based on the following considerations: (1) Global decay is unnecessary, thus some components in position encoding could be removed.
(2) Components with U-shape fluctuations within the training length lead to shortcut learning and poor extrapolation. (3) Components with lower frequencies tend to learn semantics but are not well learned. 
Since both types of components belong to the low-frequency and are mostly controlled by the latter part of the $R_{\Theta, m}$ matrix in the original RoPE, we implement our approach by replacing these components with position-independent ones while retaining the high-frequency components. We call our method \textbf{H}igh-frequency r\textbf{o}tary \textbf{P}osition \textbf{E}ncoding (\name).

We first identify the ``doomed-to-be-activated'' components and top low-frequency components in the original RoPE. As mentioned in Section~\ref{para:rope-result}, the frequencies ($\Theta_{al}$) and the minimum index $a$ of these two components  could be calculated based on the training context length $L$. The process is as follows:
\begin{equation}
\begin{aligned}
\Theta_{al} &= \{ \theta | \theta < \frac{2\pi}{L} \}, \theta \in \Theta\\
& a = \texttt{argmax}(\Theta_{al})
\end{aligned}
\end{equation}
Next, we divide the query (or key) into two parts based on the index $a$, applying positional encoding only to the first part. For the $R_{\Theta_h, m}$ matrix applied in positional encoding, we obtain it by setting $\Theta_{h} = \Theta - \Theta_{al}$. The entire process is shown in the following formula.
\begin{equation}
\begin{aligned}
& q_{m, h} = q_{m}[ :2a], k_{n, h} = k_{n}[ :2a]\\
&q_{m, h} = R_{\Theta_{h}, m}q_{h}, \quad k_{n, h} = R_{\Theta_{h}, n}k_{h},\\
&R_{\Theta_h, m} = \texttt{Diag}( R_{\theta_0, m},  ...,  R_{\theta_{a-1}, m} )\\
&q_{m} \cdot k_{n} = q_{h}^{\top} R_{\Theta_{h}, m-n} k_{h} + q_{l}^{\top} k_{l} 
\end{aligned}
\end{equation}

\subsection{Effect Verification of \name}
\label{sec:search}
\subsubsection{Evaluation}
Traditional methods for testing extrapolation usually use perplexity (PPL) as a metric. However, previous studies~\cite{Press2021TrainST} indicate that perplexity (PPL) does not effectively reflect a model's ability to fully leverage the context. In fact, a model can achieve lower PPL by primarily focusing on nearby tokens within the training length. Therefore, to more comprehensively assess the model's extrapolation, along with its contextual awareness and instruction-following potential, we additionally design two simple tasks: copying and few-shot learning.

\paragraph{Perplexity} Perplexity (PPL) is a commonly used metric for evaluating a model's extrapolation capability. We conduct our evaluation on a subset of the C4 dataset~\cite{2019t5} with 1,000 samples by comparing the zero-shot perplexity of the last 256 tokens across different input lengths.

\begin{figure}[t]
    \centering
    \includegraphics[width=0.6\linewidth,height=2.1cm]{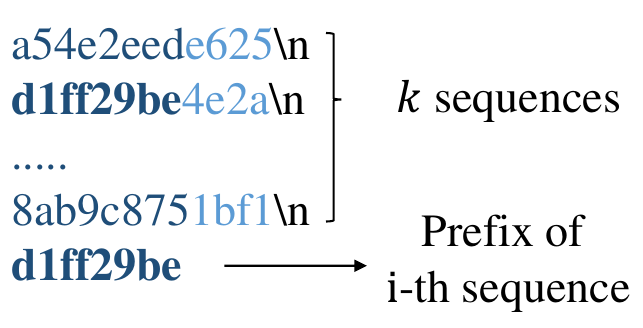}
    \caption{Specific input example of copying task.}
    \label{fig:copy}
\end{figure}

\paragraph{In-Context Copying}
The copying capability is one of the most fundamental abilities of language models and is closely related to token order. Many previous works~\cite{Liu2023LostIT, Golovneva2024ContextualPE, Lv2024LanguageMC} on model structure optimization have designed similar tasks to evaluate the models' effectiveness. Based on these studies, We designed our copying task. Specifically, we constructed a test set containing 500 samples, with each sample consisting of multiple sequences. Each sequence has an average length of 12 tokens, with a unique 8-gram prefix and 4-gram suffix. During testing, we concatenate a specific number of sequences with the prefix of a certain $i$-th sequence (queried sequence) to serve as the model’s input. The model’s objective is to output the suffix of the queried prefix, with the middle sequence selected as the queried one. Figure~\ref{fig:copy}  illustrates a specific example input case.

\paragraph{Few-shot Following}
Few-shot learning is another core ability of the model and serves as the foundation for instruction following. We created a test set with 600 samples selected from three tasks (SST-2, QNLI and RTE) in the GLUE~\cite{Wang2018GLUEAM} benchmark. For each input, we concatenate few-shot examples, a set of meaningless sentences, and the queried input. Specific examples can be found in Figure~\ref{fig:app-few}. As for the evaluation metric, instead of focusing on actual accuracy, we emphasize whether the model's output falls into the label sets from the few-shot examples. For instance, if the label set in the contextual examples is ${0, 1}$, the model's output, whether 0 or 1, will be counted. And we define this as a measure of follow ability (FA). We present the average performance across the three tasks in the main text, and detailed results for each task please refer to Appendix~\ref{app:few-shot}.

We set the rotary base $b=10,000$ in \name. Other settings are consistent with those in Section~\ref{sec:setting}. 
We evaluate the proposed \name\ against a range of established baselines, including RoPE~\cite{Su2021RoFormerET}, ALiBi~\cite{Press2021TrainST}, KERPLE~\cite{Chi2022KERPLEKR}, and FIRE~\cite{Li2024FunctionalIF}, as well as two typical RoPE-based extrapolation methods: PI~\cite{Chen2023ExtendingCW} and YaRN~\cite{Peng2023YaRNEC}.

\subsubsection{Results}

\begin{figure}[t]
    \centering
    \includegraphics[width=\linewidth,height=6cm]{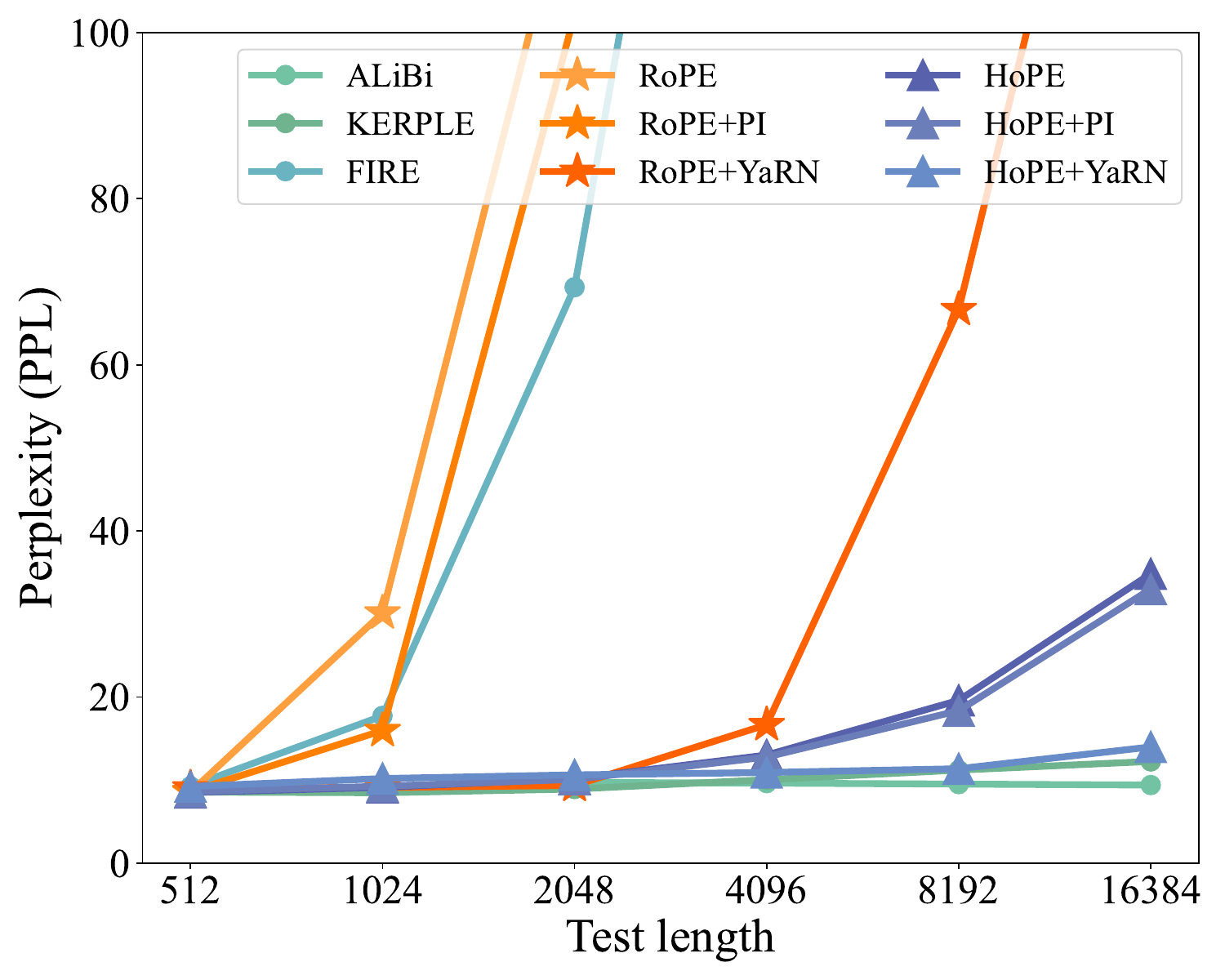}
    \caption{Perplexity Comparison on C4 dataset.}
    \label{fig:ppl}
\end{figure}

The results for perplexity (PPL), copying task, and few-shot following task are presented in Figure~\ref{fig:ppl}, Table~\ref{tab:copy} and Table~\ref{tab:few-shot}, respectively. From these results, we can draw the following conclusions:

(1) \textbf{From all perspectives in the figure and tables above, it can be confirmed that our approach significantly enhances the context awareness and extrapolation of the original RoPE.} As shown in Figure~\ref{fig:ppl}, our method noticeably smooths the increase in PPL observed in RoPE, achieving low PPL even with training lengths 4 times longer or more. It records a PPL of 8.5241 at 512, and 13.0257 at 4,096. Table~\ref{tab:copy} and~\ref{tab:few-shot}  further demonstrate that our approach not only improves extrapolation but also enhances context awareness within the training length. Specifically, compared to RoPE, the model's copy ability increased from an average of 23.80 to 60.23, while its few-shot following capability improved from an average of 54.10 to 79.20.

(2) \textbf{When combined with extrapolation methods like PI and YaRN, our method achieves even better extrapolation results.}  As shown in Table~\ref{tab:copy} and~\ref{tab:few-shot}, our method, when integrated with YaRN, achieves the best overall performance across all PEs, with an average score of 68.63 in the copy task and 93.77 in the few-shot following task. However, as also noted in the tables, while both PI and YaRN enhance extrapolation, they appear to negatively impact the model's context awareness within the training length.

\begin{table}[t]
\centering
\resizebox{\linewidth}{20mm}{
\begin{tabular}{cccccccc}\toprule
\multirow{2}{*}{Method} & \multicolumn{7}{c}{Sequence Number}\\
         & 30    & 40    & 50    & 60    & 70    & 80    & Avg.  \\ \hline
ALiBi    & 78.00 & 64.00 & 25.90 & 8.30  & 2.30  & 1.20  & 29.95 \\
KERPLE   & 80.20 & 74.00 & 64.60 & 56.80 & 25.60 & 21.70 & 53.82 \\
FIRE     & 70.00 & 43.20 & 28.20 & 6.40  & 2.80  & 0.20  & 25.13 \\ \hline
RoPE     & 77.60 & 55.80 & 9.40  & 0.00  & 0.00  & 0.00  & 23.80 \\
+PI      & 76.40 & 61.40 & 20.20 & 4.60  & 0.00  & 0.00  & 27.10 \\
+YaRN    & 65.20 & 49.80 & 64.40 & 50.80 & 48.60 & 39.40 & 54.20 \\ \hline
Our \name &  \textbf{84.00} &  \textbf{77.00} &  \textbf{77.00} & 60.40 & 32.40 & 30.60 & 60.23 \\
+PI      & 80.20 & 74.20 & 73.60 & 60.60 & 36.80 & 39.40 & 60.80 \\
+YaRN    & 78.60 & 73.20 & 76.00 &  \textbf{69.00} &  \textbf{68.40} &  \textbf{46.60} & \textbf{68.63}  \\\bottomrule 
\end{tabular}
}
\caption{ Results of the copy task. We highlight the leading results with \textbf{bold fonts}. The input length is under the training length of 512 when the sequence number is below 60.}
\label{tab:copy}
\end{table}

(3) \textbf{Relying solely on perplexity (PPL) to measure extrapolation is not reliable.} In some cases, the PPL measurements (shown in Figure~\ref{fig:ppl}) contradict the performance in other tasks (shown in Table~\ref{tab:copy} and~\ref{tab:few-shot}), indicating that PPL may not prove a method's ability to effectively utilize global information. It might reflect ``pseudo'' extrapolation, as seen in the result of ALiBi and KERPLE. From Table~\ref{tab:copy} and~\ref{tab:few-shot}, it is evident that ALiBi's actual extrapolation is poor, and while KERPLE shows some extrapolation, it is not as strong as the PPL suggests and performs slightly worse in few-shot following performance.

\begin{table}[t]
    \centering
    \resizebox{\linewidth}{!}{
    \begin{tabular}{ccccccc}
   \toprule
   \multirow{2}{*}{Method} & \multicolumn{6}{c}{Input Lengths}\\
   & 256   & 512   & 768   & 1024  & 1280  & Avg.  \\ \hline
ALiBi    & 99.67 & 85.67 & 68.67 & 6.33  & 1.00  & 52.27 \\
KERPLE   & 77.17 & 70.17 & 22.33 & 14.67 & 16.33 & 40.13 \\
FIRE     & 87.17 & 86.33 & 32.17 & 19.50 & 19.33 & 48.90 \\ \hline
RoPE     &  98.17 & 97.00 & 51.33 & 17.00 & 7.00  & 54.10 \\
\textit{+PI}       & 98.17 &  \textbf{98.83} & 66.67 & 38.50 & 8.00  & 62.03 \\
\textit{+YaRN}        &  \textbf{99.83} & 85.67 & 91.00 & 81.67 & 20.33 & 75.70 \\ \hline
Our \name & 99.67 & 98.33 & 74.50 & 67.83 & 55.67 & 79.20 \\
\textit{+PI}       & 99.17 & 98.50 & 83.67 & 78.67 & 51.67 & 82.33 \\
\textit{+YaRN}     & 97.33 & 92.17 & \textbf{91.67} & \textbf{88.67} & \textbf{99.00} & \textbf{93.77}
\\\bottomrule
    \end{tabular}}
    \caption{The results of the few-shot following experiment. We measure the model's following ability (FA), which counts the instances when the output includes one of the label sets from the examples. The leading results are highlighted with \textbf{bold fonts}.}
    \label{tab:few-shot}
\end{table}

\subsection{Attention Patterns in \name}
\label{app:pattern-hope}
\begin{figure}[!ht]
 \includegraphics[width=\linewidth]{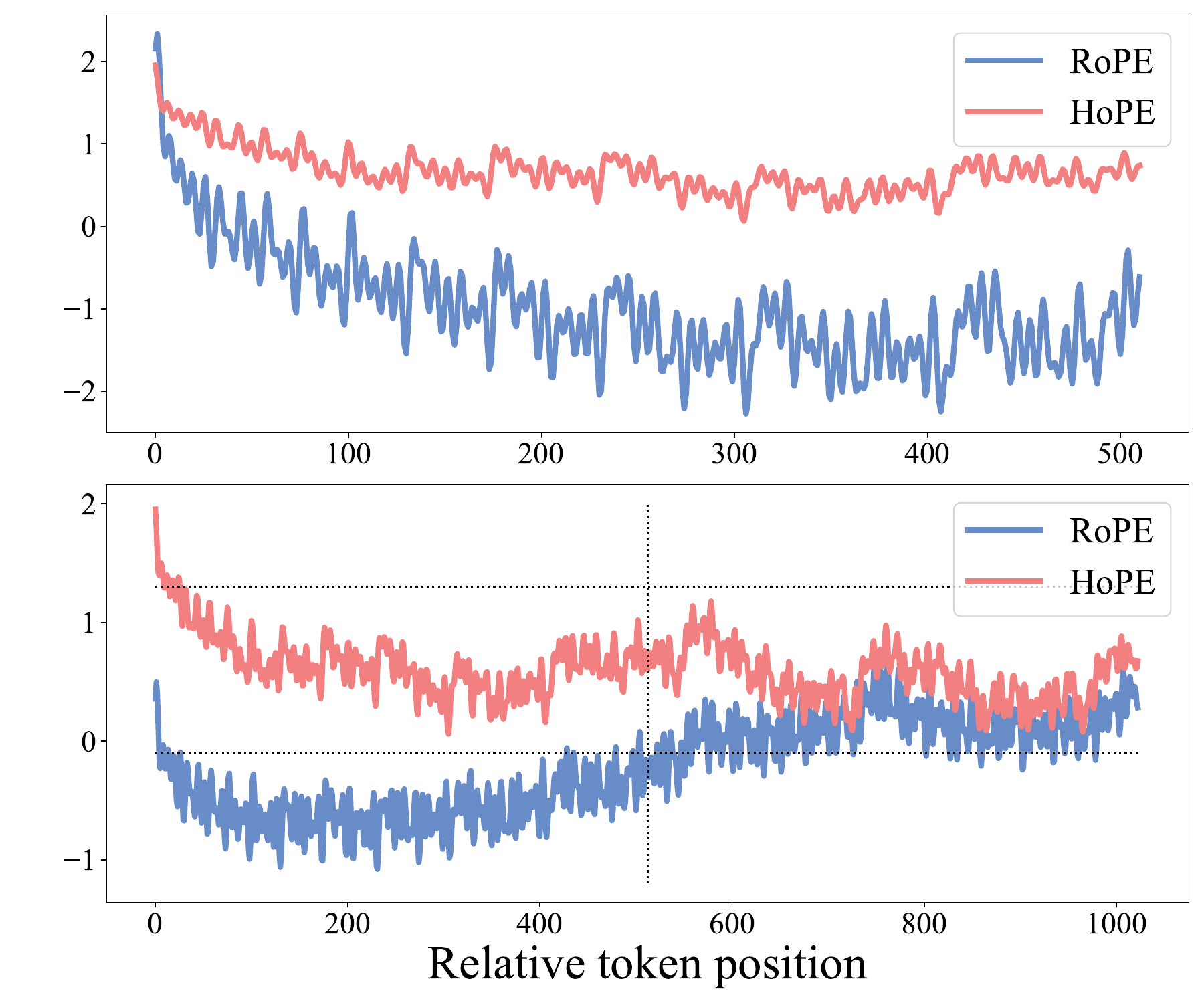}
\caption{Attention patterns in \name\ and RoPE, both within (top) and beyond (bottom) training length.}
\label{fig:app-hope}
\end{figure}

We provide a observation about the position-related attention patterns of \name\ in Figure~\ref{fig:app-hope}. As shown in the upper subplot, \name\ demonstrates a U-shaped fluctuation similar to RoPE within the training length. However, the positional fluctuation in \name\ is somewhat milder, indicating that \name\ may be better adapted to long-context tasks which requires greater focus on semantic information. Regarding the extrapolation patterns (shown in the lower subplot of Figure~\ref{fig:app-hope}), \name\ does not exhibit the out-of-distribution (OOD) behavior seen in RoPE, which could account for its superior extrapolation capabilities.

\section{Ablation Study of \name}

In this section, we conducted an ablation study of our method and validated its effectiveness through measurements on the copy task. 
\paragraph{Settings}We mainly performed three ablation settings: 
\begin{itemize}
    \item \textbf{AB1}: replacing only the ``activated'' components with positional-independent components. 
    \item \textbf{AB2}: replacing only the top low-frequency components with positional-independent components. 
    \item \textbf{AB3}: removing both components without adding positional-independent ones. We do this by replacing these components with high-frequency components.
\end{itemize}

\paragraph{Results}
As shown in Table~\ref{tab:ab}, 
removing the ``activated'' components (AB1) results in a significant improvement in both context awareness and extrapolation, with an average increase of 14.5 points over RoPE. This outcome suggests that these  components indeed contribute to the model's short-cut learning, hindering optimal learning. 

Removing the top low-frequency components (AB2) helps improve the model's context awareness but contributes less to extrapolation. This confirms that the ``activated'' components are the key factor behind poor extrapolation performance.  

Additionally, we can observe that AB3 shows a significant decline in both context awareness and extrapolation, highlighting the importance of the position-independent components. This indicates that the model indeed requires certain components to learn semantic information. The slight improvement (an average increase of 0.67 points) from AB2 further suggests that the original low-frequency components in RoPE effectively fulfill this role, while they have not been fully learned.

Based on the results above, we have demonstrated the rationale behind our \name's design and identified the source of its performance improvements.

\begin{table}[t]
\centering
\resizebox{\linewidth}{14mm}{
\begin{tabular}{cccccccc}\toprule
\multirow{2}{*}{Method} & \multicolumn{7}{c}{Sequence Number}\\
         & 30    & 40    & 50    & 60    & 70    & 80    & Avg.  \\ \hline
RoPE    & 77.60 & 55.80 & 9.40  & 0.00  & 0.00  & 0.00  & 23.80 \\
AB1   & 83.80 & 68.80 & 17.80   & 21.80 & 12.00 & 25.60 & 38.30 \\
AB2      & 81.40 &  59.60 & 16.80  &  1.00  &  0.00   &  0.00 &24.47 \\
AB3      & 65.60 & 30.6 & 6.20  & 2.00  & 0.80  & 0.00  & 17.53 \\
\name &  \textbf{84.00} &  \textbf{77.00} &  \textbf{77.00} & \textbf{60.40} & \textbf{32.40} & \textbf{30.60} & \textbf{60.23} 
\\ \bottomrule 
\end{tabular}
}
\caption{The results of the ablation study. AB1 means only removing the ``activated'' components, AB2 means removing the top low-frequency components, and AB3 refers to removing both types of components but without the position-independent components.}
\label{tab:ab}
\end{table}

\section{Scalability of \name}
Based on our understanding of \name's advantages, derived from a series of empirical experiments with small models and toy tasks, we conducted further comparative experiments on real-world tasks. 
These experiments involved training large language models from scratch, comparing RoPE with \name.

\paragraph{Settings} We trained a 3B Llama-based model with a training length of 8192 over 120k steps, processing approximately 500B tokens in total. The model configurations and training details can be found in Appendix~\ref{app:setting}.
For evaluation, we selected 8 general-purpose benchmarks, including MMLU(5-shot)~\cite{Hendrycks2020MeasuringMM}, MMLU-PRO(5-shot)~\cite{Wang2024MMLUProAM}, GPQA(0-shot)~\cite{Rein2023GPQAAG}, BBH(3-shot)~\cite{Srivastava2022BeyondTI}, WinoGrande(5-shot)~\cite{Sakaguchi2019WinoGrande}, GSM8k(8-shot)~\cite{Cobbe2021TrainingVT}, MATH(4-shot)~\cite{Lightman2023LetsVS} and DROP(3-shot)~\cite{Dua2019DROPAR}.  These datasets allow for a comprehensive evaluation of the model's performance across various tasks such as multiple-choice questions, reasoning, and numerical computations. The shot count settings follow several works~\cite{Dubey2024TheL3, Bai2023QwenTR} and we use OpenCompass~\cite{Fu2024GETAF} to compute the results.

\paragraph{Results}
As shown in Table~\ref{tab:3b}, our \name\ achieves better overall performance compared to RoPE, with a higher average score (26.42) and notable improvements on tasks like MMLU, GPQA, and DROP.  We also provide the results of each dataset for every 100B token updates in Appendix~\ref{app:3b}, which further illustrate \name's consistent superiority across most datasets and highlight its robust advantages.
These findings highlight the potential of \name\ to drive the next generation of state-of-the-art LLMs by replacing the widely used RoPE.

\section{Conclusion}
In this paper, we explore the empirical attention patterns of various positional encodings and observe that position-related attention tend to form a U-shape pattern, benefiting more from local decay rather than global. Our further analysis of RoPE reveals a strong correlation between the U-shape pattern and its learned components. We identify that certain "activated" components and top low-frequency components in RoPE hinder the model's optimal learning process, limiting its context awareness and extrapolation.  Consequently, we propose our method, \name, which breaks the principle of long-term decay in theory, allowing for optimal utilization of components for positional encoding. Extensive experiments demonstrate its effectiveness in enhancing both context awareness and extrapolation.

\bibliography{anthology,custom}

\appendix
\newpage

\section{Model Configuration}
\label{app:setting}

Detail settings across model sizes are depicted in Table~\ref{tab:setting}. Other hyperparameters are as follows: the AdamW optimizer is used with $(\beta_1, \beta_2) = (0.9,0.999)$, a learning rate of $3e^{-4}$, $2,000$ warm-up steps, and a gradient clipping value of 1. Experiments for the 125M models are conducted on 8 A100 GPUs, while those for the 3B models use 256 A100 GPUs. 
\begin{table}[ht]
    \centering
    \resizebox{0.9\linewidth}{!}{
    \begin{tabular}{lcc}
        \toprule
         Hyperparameters     &  	125M & 3B \\
        \hline
       Training sequence length & 512& 8192\\
        Batch size & $64 \times 8$  & $2 \times 256$ \\
       Number of Iterations & 50k & 120K\\
        Number of Layers & 12 & 34 \\
        Attention Head & 12 & 16 \\
        Feature Dimension & 768 & 2048  \\
        Intermediate Dimension & 2688 & 8704\\
       Precision & Float16 & Float16 \\
        \bottomrule
    \end{tabular}}
    \caption{Model configurations.}
    \label{tab:setting}
\end{table}

\section{Supplementary Results on the Exploration of Component Effects in RoPE Attention Patterns}
\label{app:pattern-rope}

\begin{figure}[ht]
    \centering

    \includegraphics[width=\linewidth]{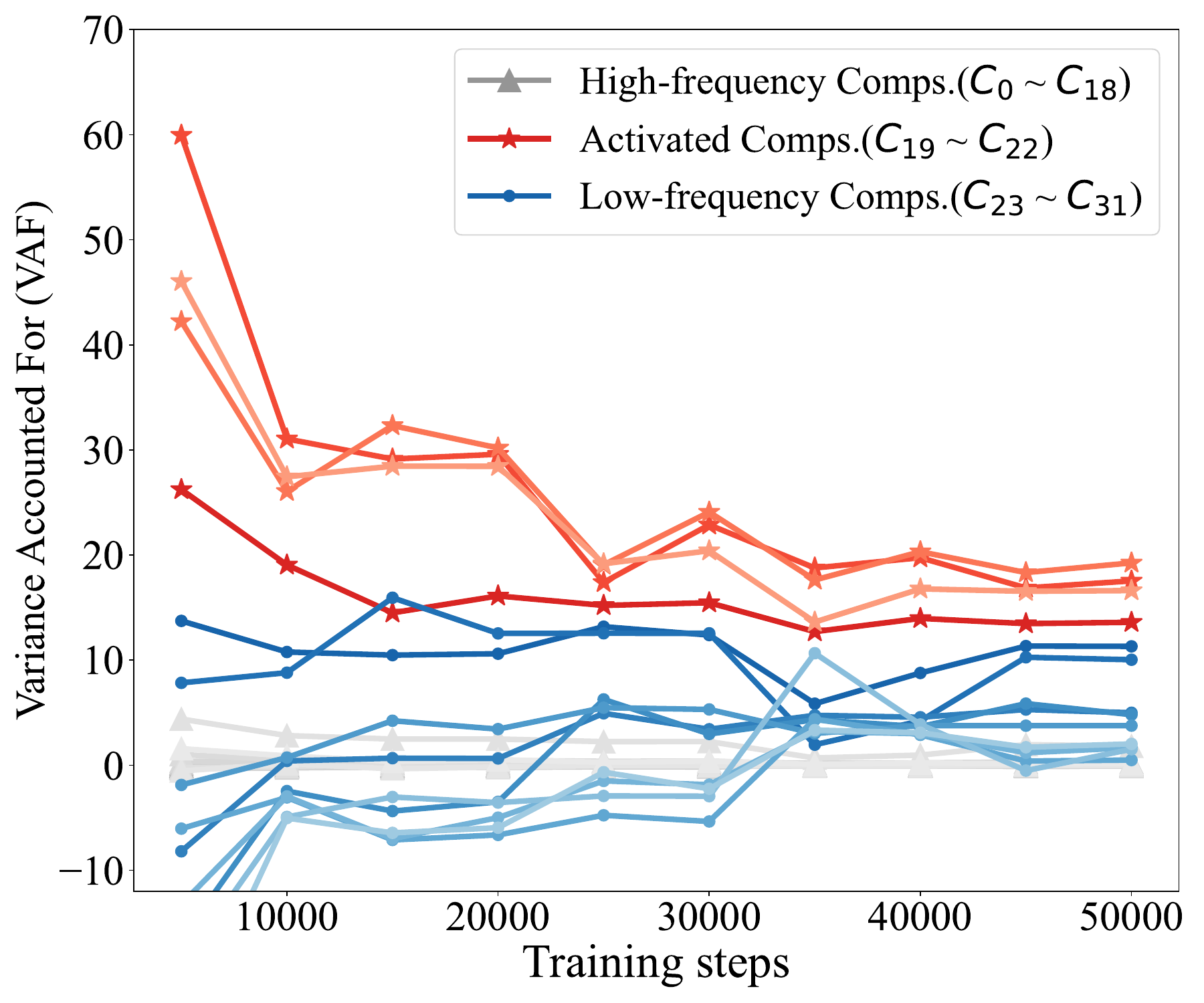}

    \caption{VAF results of each component at training length 1,024.} 
    \label{fig:pattern-rope-app-a}
\end{figure}
\begin{figure}[ht]
    \centering
    \includegraphics[width=\linewidth]{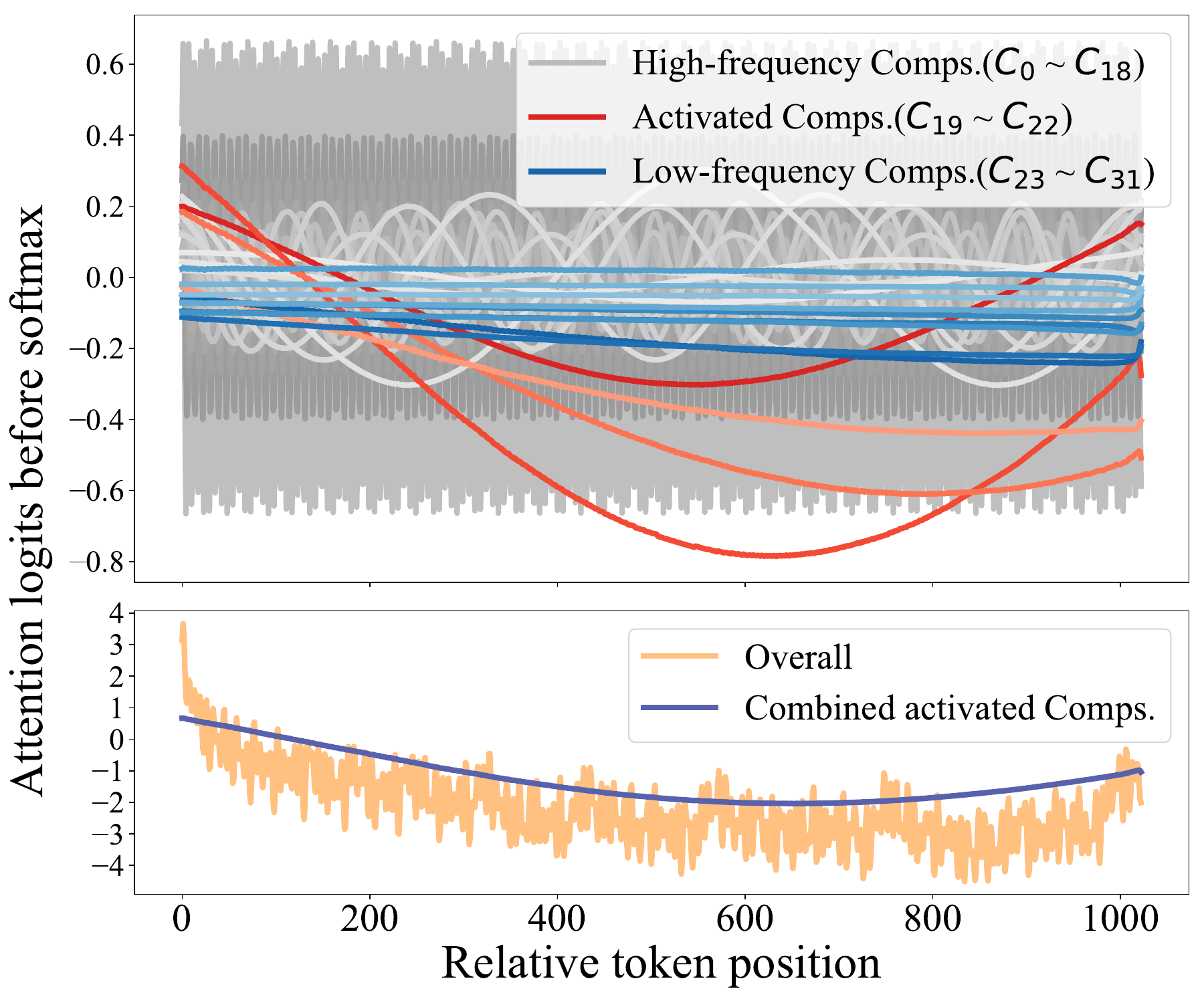}
    \caption{The attention pattern of RoPE at training length 1,024.}
    \label{fig:pattern-rope-app-b}
\end{figure}

We present supplementary experiments with training lengths of 1024, depicted in Figure~\ref{fig:pattern-rope-app-a} and ~\ref{fig:pattern-rope-app-b}. We reached the same conclusion as in the main text. It can be seen that ``activated'' and lower frequency components shift further back as the training length increases. These ``activated'' components still exhibit a U-shaped curve, similar to the final pattern.  The lower frequency components continue to learn a constant pattern. 

\section{Supplementary for Few-shot Following Tasks}

\subsection{Input Examples of Few-shot Following Tasks}
\label{app:few-shot-example}

\begin{figure}[!ht]
 \includegraphics[width=\linewidth]{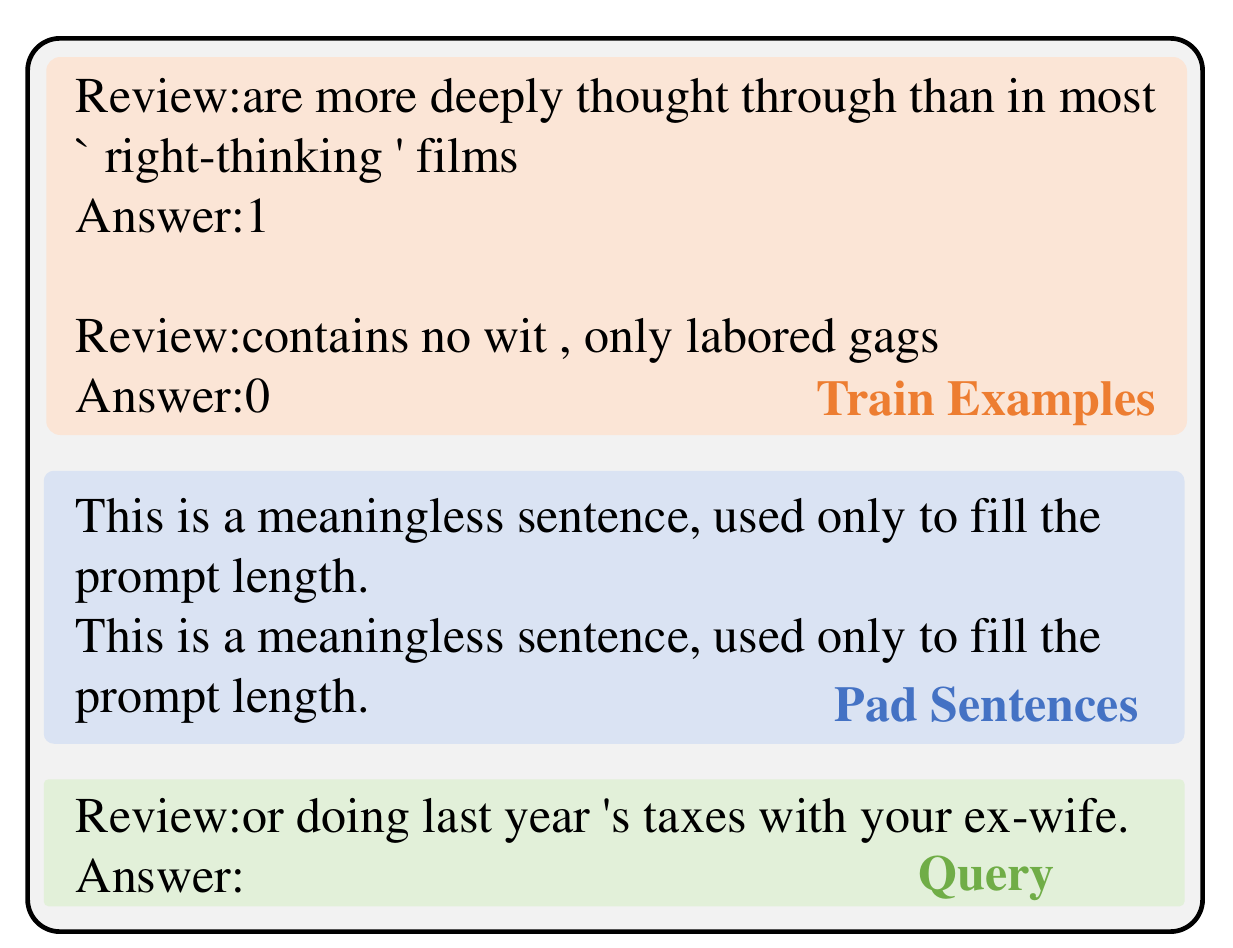}
\caption{Input examples of few-shot following tasks. We take SST-2 as an instance.}
\label{fig:app-few}
\end{figure}

Specific example of few-shot following tasks can be found in Figure~\ref{fig:app-few}. In practice, We provide 5-shot training examples as context. For each instance, we dynamically pad it with a different number of meaningless sentences to ensure the various input lengths.

\subsection{Detail Results on Few-shot Following Task}
\label{app:few-shot}
We present the average results in Table~\ref{tab:few-shot} of  main text.  Detail results in three tasks are depicted in Table~\ref{tab:app-few}. 

\begin{table*}[ht]
\resizebox{\linewidth}{!}{
\begin{tabular}{c|cccccc|cccccc|cccccc}
\toprule
\multirow{2}{*}{Method} & \multicolumn{6}{c|}{SST-2}                        & \multicolumn{6}{c|}{QNLI}                             & \multicolumn{6}{c}{RTE}                            \\
                   & 256    & 512   & 768   & 1024  & 1280  & Avg.  & 256    & 512    & 768    & 1024   & 1280   & Avg.  & 256    & 512    & 768    & 1024  & 1280  & Avg.  \\ \hline
ALiBi              & 99.00  & 58.00 & 29.00 & 4.00  & 0.00  & 38.00 & 100.00 & 99.00  & 79.00  & 0.00   & 0.00   & 55.60 & 100.00 & 100.00 & 98.00  & 15.00 & 3.00  & 63.20 \\
KERPLE             & 50.50  & 37.50 & 18.00 & 9.00  & 6.00  & 24.20 & 100.00 & 95.00  & 32.00  & 7.00   & 4.00   & 47.60 & 81.00  & 78.00  & 17.00  & 28.00 & 39.00 & 48.60 \\
FIRE               & 99.50  & 99.00 & 76.50 & 31.50 & 3.00  & 61.90 & 99.00  & 99.00  & 18.00  & 19.00  & 12.00  & 49.40 & 63.00  & 61.00  & 2.00   & 8.00  & 43.00 & 35.40 \\ \hline
RoPE               & 99.50  & 96.00 & 20.00 & 0.00  & 0.00  & 43.10 & 95.00 & 95.00 & 85.00  & 22.00  & 9.00   & 61.20 & 100.00 & 100.00 & 49.00  & 29.00 & 12.00 & 58.00 \\
+ PI               & 99.50  & 96.50 & 33.00 & 4.50  & 0.00  & 46.70 & 100.00 & 100.00 & 94.00  & 75.00  & 11.00  & 76.00 & 95.00 & 100.00 & 73.00  & 36.00 & 13.00 & 63.40 \\
+ YaRN             & 99.50  & 57.00 & 73.00 & 73.00 & 0.00  & 60.50 & 100.00 & 100.00 & 100.00 & 95.00  & 9.00   & 80.80 & 100.00 & 100.00 & 100.00 & 77.00 & 52.00 & 85.80 \\ \hline
Our \name           & 99.00  & 95.00 & 45.50 & 52.50 & 32.00 & 64.80 &100.00  & 100.00  & 100.00 & 88.00  & 37.00  & 85.00 & 100.00  & 100.00  & 78.00  & 63.00 & 98.00 & 87.80 \\
+ PI               & 97.50  & 97.50 & 97.00 & 37.00 & 89.00 & 83.60 & 100.00 & 100.00 & 100.00 & 100.00 & 55.00  & 91.00 & 100.00  & 98.00  & 54.00  & 99.00 & 11.00 & 72.40 \\
+ YaRN             & 100.00 & 97.50 & 76.00 & 69.00 & 98.00 & 88.10 & 92.00  & 80.00  & 99.00  & 99.00  & 100.00 & 94.00 & 100.00 & 99.00  & 100.00 & 98.00 & 99.00 & 99.20
                 
   \\
\bottomrule
\end{tabular}}
\caption{Detail results on three few-shot tasks.}
\label{tab:app-few}
\end{table*}

\begin{figure*}[ht]
\includegraphics[width=\linewidth]{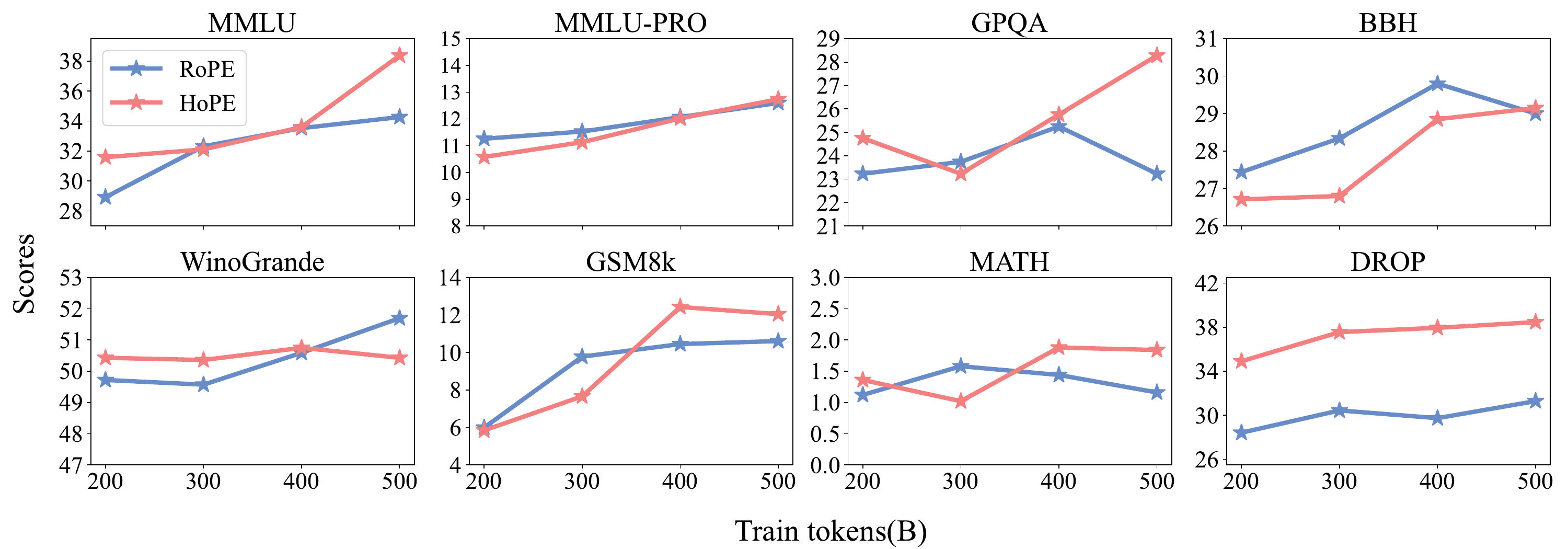}
\caption{Performance comparison across 8 benchmarks using a 3B Llama-based model with different  number of tokens trained.}
\label{fig:app-3b}
\end{figure*}

\section{Supplementary Results of the 3B Llama-based Model}
\label{app:3b}
In this section, we provide the supplementary results of the 3B Llama-based Model with a training length 8192. As depicted in Figure~\ref{fig:app-3b}, we evaluate the results on each dataset every 100B token updates. 

\end{document}